\documentclass[10pt,twocolumn,letterpaper]{article}

\usepackage{wacv}
\usepackage{times}
\usepackage{epsfig}
\usepackage{graphicx}
\usepackage{amsmath}
\usepackage{amssymb}
\usepackage{booktabs}
\usepackage{multirow}
 
\def\wacvPaperID{2} 

\wacvalgorithmstrack   

\wacvfinalcopy 

\ifwacvfinal
\usepackage[breaklinks=true,bookmarks=false]{hyperref}
\else
\usepackage[pagebackref=true,breaklinks=true,colorlinks,bookmarks=false]{hyperref}
\fi

\begin{document}

\title{Face Forgery Detection Based on Facial Region Displacement Trajectory Series}

\author{YuYang Sun$^{1,3}$, ZhiYong Zhang$^3$, Isao Echizen$^{1,2}$,
Huy H. Nguyen$^2$, ChangZhen Qiu$^3$, and Lu Sun$^4$ \\
$^1$The University of Tokyo, Japan\ \ \ \ \ \ \ \ \ \ \ \ $^2$National Institute of Informatics, Japan  \\
$^3$Sun Yat-sen University, China\ \ \ \ \ \ \ \ \ \ \ \ $^4$South China University of Technology, China \\
{\tt\small tarrysun0115@g.ecc.u-tokyo.ac.jp, \{nhhuy,iechizen\}@nii.ac.jp}
}

\maketitle
\thispagestyle{empty}

\begin{abstract}
   Deep-learning-based technologies such as deepfakes ones have been attracting widespread attention in both society and academia, particularly ones used to synthesize forged face images. These automatic and professional-skill-free face manipulation technologies can be used to replace the face in an original image or video with any target object while maintaining the expression and demeanor. Since human faces are closely related to identity characteristics, maliciously disseminated identity manipulated videos could trigger a crisis of public trust in the media and could even have serious political, social, and legal implications. To effectively detect manipulated videos, we focus on the position offset in the face blending process, resulting from the forced affine transformation of the normalized forged face. We introduce a method for detecting manipulated videos that is based on the trajectory of the facial region displacement. Specifically, we develop a virtual-anchor-based method for extracting the facial trajectory, which can robustly represent displacement information. This information was used to construct a network for exposing multidimensional artifacts in the trajectory sequences of manipulated videos that is based on dual-stream spatial-temporal graph attention and a gated recurrent unit backbone. Testing of our method on various manipulation datasets demonstrated that its accuracy and generalization ability is competitive with that of the leading detection methods.
\end{abstract}

\section{Introduction}
With the development and spread of high-resolution, and rich-diversity deep generation models such as generative adversarial networks  (GANs)~\cite{goodfellow2014generative},  images can be synthesized with sufficient texture details to fool the casual viewer. This not only illustrates the potential creativity of artificial intelligence but also lays bare the hidden dangers for digital information security. Deep-learning-based digital face manipulation technologies have gradually become widely used. These manipulation technologies can be used to maliciously edit, deform and replace faces in images or videos, which can then be used pornographically to damage reputations or to spread fake news and hate speech with a celebrity connection, causing political tensions and democratic crises. In addition, more and more open-source deepfakes applications such as FaceApp~\cite{FaceApp} and ZAO~\cite{ZAO} have reduced the need for manual editing in video face swapping, so anyone with a laptop and Internet connection can easily synthesize and spread forged face videos.

\begin{figure}[t]
	\begin{center}
		\includegraphics[width=0.9\linewidth]{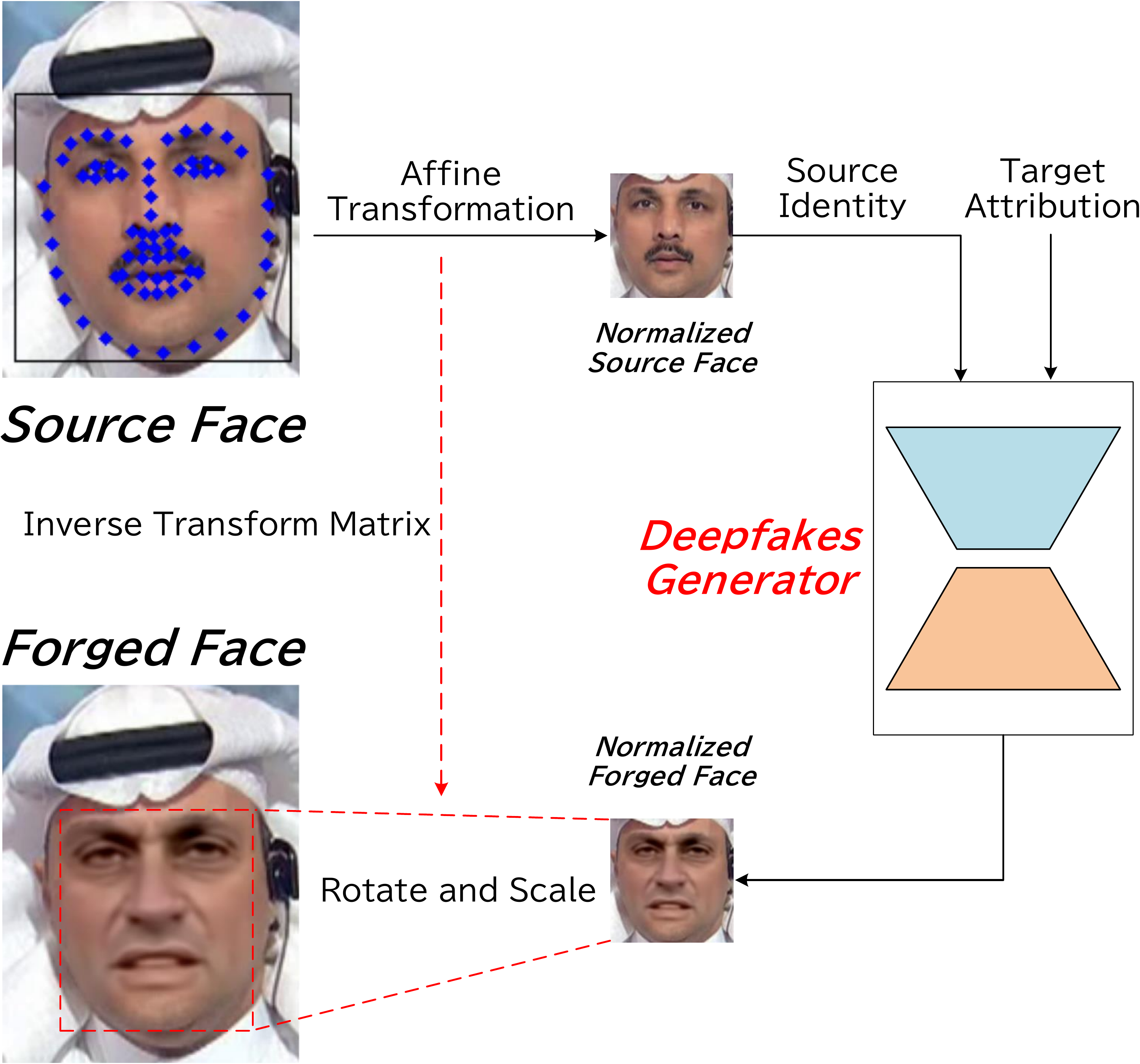}
	\end{center}
	\caption{A brief overview of DeepFake generation pipeline. The synthesized normalized manipulated face utilizes the landmark information of the source face for inverse affine transformation to match the source video frame, which may cause spatial artifacts in local facial regions and further form inconsistent temporal abruptness in frame-by-frame manipulation.}
	\label{fig:fig1}
\end{figure}
Therefore, it is essential to propose robust detection methods to counteract the potential threats caused by the proliferation of synthetic and manipulated videos in order to protect personal privacy and property security. Current mainstream image-based detection methods only focus on exposing frame-level artifacts in the spatial or frequency domain and determine the video-level results through independent multi-frame detection, which ignores the temporal inconsistency information hidden in the video stream. We assert that manipulation leaves temporally unnatural marks in several spatial regions of the face. Therefore, simultaneously capturing the intra-frame spatial coherence between facial regions and the inter-frame temporal consistency of information in specific areas is of great help in identifying forged faces. Our method is dedicated to exposing subtle spatial artifacts and inconsistent temporal abruptness in critical facial regions caused by the forced blending of the synthesized normalized forged face in the deepfakes pipeline.

As the first step in developing a robust detection method, we statistically analyze the difference in landmark sequences between real and fake videos (Section \ref{sec:3}) in order to determine whether digital face manipulation creates specific artifacts in the micro-motion trajectory of the face. Next, we propose a virtual-anchor-based method for extracting the facial region displacement trajectory (Section \ref{sec:4}). Several local facial regions are taken as tracking targets, and the feature points with excellent tracking characteristics are screened in each region frame-by-frame to robustly and accurately obtain the displacement trajectory. Finally, we develop a fake trajectory detection network (FTDN) based on dual-stream spatial-temporal graph attention and a gated recurrent unit (GRU) backbone for detecting fake trajectories (Section \ref{sec:5}).  The network utilizes the extracted trajectory and explicitly aggregates the important information of different dimensions in the input sequences, which can help to effectively capture the spatial-temporal anomalies in the manipulated video trajectory. We tested our model on various manipulation techniques in the FaceForensics++ \cite{rossler2019faceforensics++} dataset and achieved excellent detection performance. 

We summarize our contributions as follows:

\begin{itemize}
	\item We analyze and verify the impact of face forgery facial landmark series, and thus construct a DeepFakes detection method by capturing the spatial-temporal anomalies of facial region displacement trajectory;
	\item We design a robust facial region displacement trajectory extraction method based on virtual anchor to highlight the local displacement features;
	\item We propose a Fake Trajectory Detection Network (FTDN) to effectively identify forgery spatial-temporal patterns based on the facial region displacement trajectory with competitive results on the FaceForensics++ dataset.
\end{itemize}

\section{Related Work}
\subsection{Deepfakes Detection}
To identify synthetic face images, early researchers focused on capturing specific artifacts caused by defects of deepfakes generators. Well-known examples of such artifacts are inconsistent head poses~\cite{yang2019exposing}, abnormal facial expressions and head movements~\cite{agarwal2019protecting}, broken photo-response non-uniform patterns~\cite{koopman2018detection}, and detectable differences based on image quality measures~\cite{korshunov2018deepfakes}. These artifacts may strongly affect certain deepfakes methods but do not generalize well, resulting in poor robustness. Subsequent work used neural networks to adaptively mine the high-level features of forged face images and learn the pattern differences~\cite{afchar2018mesonet,nguyen2019capsule,li2018exposing}, which could substantially improve model accuracy and generalizability. For video-level detection, researchers generally believe that low-level artifacts caused by face manipulation manifest as temporal artifacts that are inconsistent across the frame, so they often take temporal information into consideration. As representative examples, Sabir \etal~\cite{sabir2019recurrent} leveraged a bidirectional GRU~\cite{cho2014learning} on the feature output of a convolutional neural network (CNN) backbone to identify the temporal patterns. Güera and Delp~\cite{guera2018deepfake} used a CNN to extract frame-level features and fed them into a long short-term memory (LSTM)~\cite{hochreiter1997long} to create temporal descriptors for classification. 

In particular, previous work~\cite{damer2018detecting,yang2019exposinggan,agarwal2019protecting,sun2021improving} has shown the potential of using facial geometric information in face manipulation detection. However, these works did not simultaneously mine the temporal inconsistency of facial geometric features caused by frame-by-frame manipulation and the spatial information anomalies between different facial regions caused by the distortion in the process of forgery face synthesis. Inspired by the previous works, we choose to transform the task of detecting deepfakes videos into the task of detecting multi-variable time series anomalies to expose artifacts caused by facial manipulation in both temporal and spatial dimensions.

\subsection{Graph Neural Networks}
Graph neural networks (GNNs) are deep learning models based on a graph structure, in which the nodes and edges are used to structure the content and attributes of the target so as to improve the representation ability of non-Euclidean data. Scarselli \etal~\cite{scarselli2008graph} first proposed the concept, i.e., the combining of graph data with a neural network to carry out end-to-end calculation on structured data. Bruna \etal subsequently proposed incorporating local feature extraction and weight sharing into the calculation of the graph structure, resulting in a graph convolutional network (GCN)~\cite{bruna2013spectral}. 

Further studies revealed that the attention mechanism can help the neural networks effectively capture the essential features of the data and avoid redundant information. This led to Veličković \etal devising the graph attention network (GAT)~\cite{velivckovic2017graph}, a special form of the spatial-based GCN. They introduced the attention mechanism into node aggregation, aiming to assign weights to neighbor nodes in accordance with their importance when aggregating the information of the target node. Brody \etal~\cite{brody2021attentive} improved the attention formula and proposed using dynamic GAT to further improve the effect of aggregation. Here we use the dynamic graph attention mechanism to explicitly add attention weights to the spatial-temporal trajectory sequence, which effectively helps our model to focus on the anomalies caused by face forgery.

\subsection{Time Series Processing}
A time series is a very common form of data and is generally a sequence of measured values obtained by sampling variables in time-series order. Traditional time-series processing uses an auto-regressive moving average model to fit the sequences. Advances in machine learning led to some researchers such as Kate \etal combining k-nearest neighbor classification with dynamic time warping~\cite{itakura1975minimum} to robustly represent misaligned time-series data~\cite{kate2016using}. Subsequent work demonstrated that deep learning methods are effective for processing time-series data. For instance, Malhotra \etal devised a pre-trained deep recurrent neural network called ``TimeNet” for classifying time-series data~\cite{malhotra2017timenet} that uses a recurrent neural network (RNN) to construct encoder-decoder pairs and to learn a pre-trained temporal feature extraction model in an unsupervised way. Karim \etal. proposed using an LSTM-FCN architecture~\cite{karim2019insights} to learn the local features and global correlation of time series data.

Some researchers have speculated that the correlation between features and between time steps in a multi-variable time series can be effectively represented by a graph structure and have combined GNNs with time series learning. For example, Zhao \etal proposed using a MTAD-GAT~\cite{zhao2020multivariate}, which combines graph attention information into a time-series representation, to effectively highlight the anomalies in sequences. Similarly, Deng and Hooi~\cite{deng2021graph} created a learnable graph structure that enables the time-series anomaly detection model to adaptively learn the relationships between feature nodes. Inspired by these ideas, we decided to use dual-stream spatial-temporal graph attention to identify as much as possible the abnormal artifacts in the trajectory sequences of manipulated videos.


\begin{figure}[t]
	\begin{center}
		\includegraphics[width=1.01\linewidth]{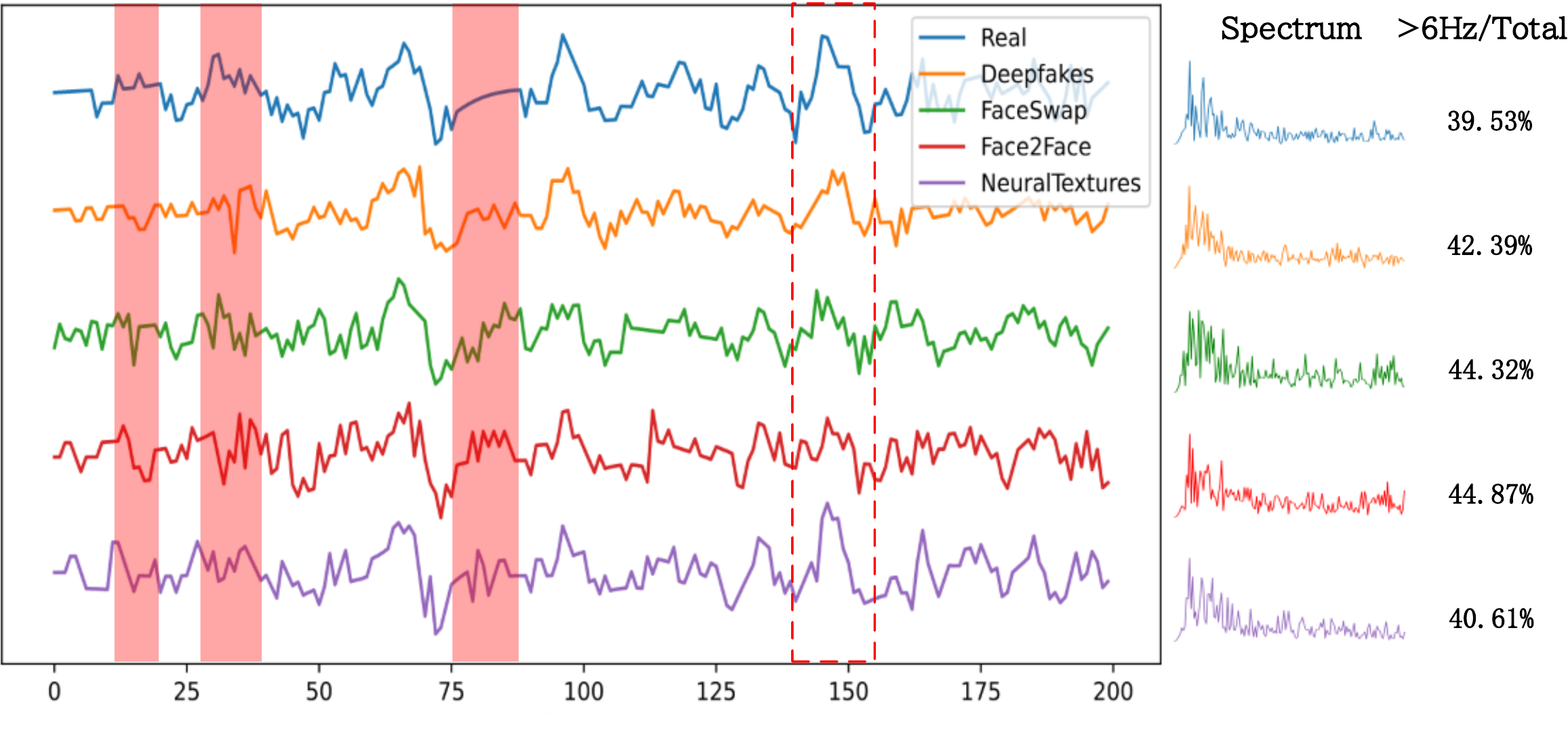}
	\end{center}
	\caption{Landmark longitudinal trajectory diagram and corresponding spectrum for real video and several manipulated videos (series have been normalized and detrended).}
	\label{fig:fig2}
\end{figure}

\begin{figure*}
	\begin{center}
		\includegraphics[width=0.85\linewidth]{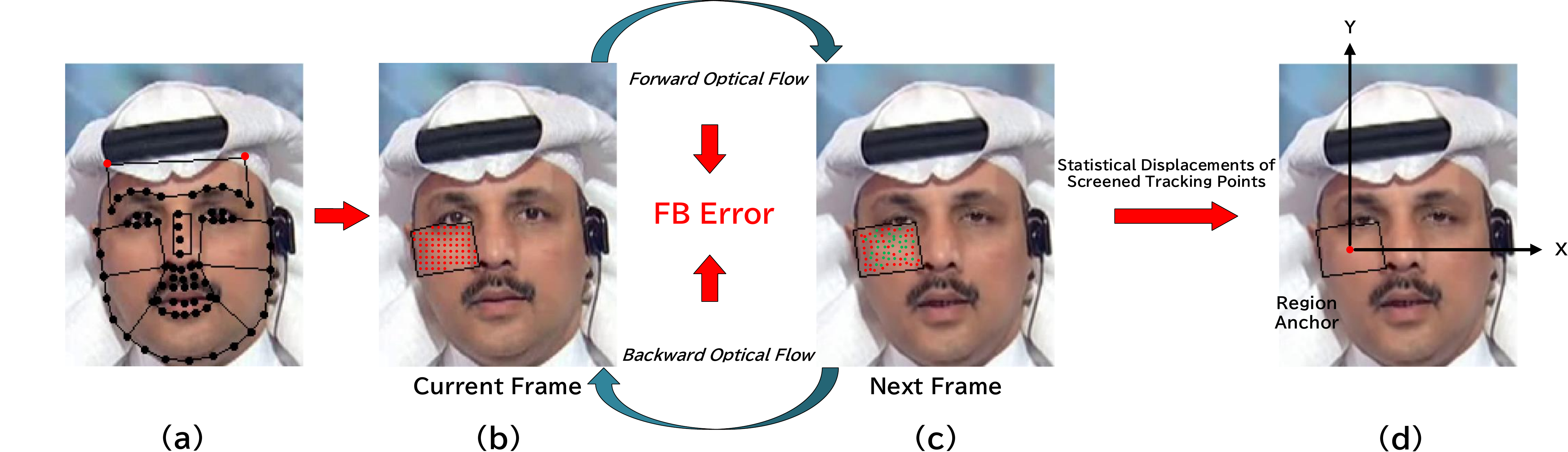}
	\end{center}
	\caption{Overview of proposed virtual-anchor-based facial region displacement trajectory extraction method. (a) For each frame, landmarks are first detected and seven ROIs are defined; (b) Tracking points are marked at uniform intervals in each ROI; (c) forward and backward optical flows between current and next frame for tracking points are calculated to obtain forward-backward (FB) errors; (d) points with excellent tracking characteristics (green points) are screened by FB error and statistical displacements of these points are used to represent displacements of virtual anchors in ROIs}
	\label{fig:fig3}
\end{figure*}

\section{Method Feasibility Analysis}
\label{sec:3}
Since a deepfakes generator requires input and output images of fixed size, the source face needs to be normalized. As shown in Figure~\ref{fig:fig1}, the landmarks in the source face are first extracted, and these feature coordinates are used to construct an affine transformation matrix, which is used to convert the source face into a normalized form with fixed size and head pose. The deepfakes generator then combines the identity information of the source face with the attribute information of the target face to synthesize a forged face. To ensure that the forged face perfectly matches the contour of the source face, the above matrix is used for inverse affine transformation of the normalized forged face. Analysis of this process led to the following conjecture: Due to the structural differences from person to person, using the source face's features to forcibly rotate and scale the manipulated output inevitably introduces spatial errors into local regions of the forged face, and further become capturable temporal artifacts in trajectory sequence during frame-by-frame processing. These artifacts could be helpful in deepfake detection.

To prove our conjecture, we extracted the temporal trajectories of the longitudinal position of a specific landmark in an original video and several corresponding manipulated versions. As shown in Figure~\ref{fig:fig2}, the trends of these trajectories are generally consistent; however, those for the forgeries are completely opposite to that for the real trajectory at three time steps (red-shaded areas). In addition, the trajectories of the forgeries have more burrs and spikes and are much coarser than that of the real trajectory (e.g., red-outlined box). The figure also shows the spectra of these trajectories and the proportion of high-frequency components ($>6 Hz/Total$). The real trajectory had the smallest proportion of high-frequency components (39.53\%). The Deepfakes, FaceSwap, and Face2Face forgeries obviously had more high-frequency burr noise and higher proportions (42.39\%, 44.32\%, and 44.87\%, respectively). Due to the textures rendering, NeuralTextures had a smaller fusion error, so its trajectory is similar to that of the real one, and its proportion of high-frequency components relatively low.

The high-frequency burrs are attributed to the spatial error caused by the abnormal warping of the forged face during the inverse affine transformation and are reflected in the abnormal movement trends of the feature points in the trajectories. This suggests that the facial region displacement trajectory is a robust feature with pattern differences and that its spatial-temporal information can be used to effectively expose the face manipulation.


\section{Trajectory Extraction Method}
\label{sec:4}
Existing methods for extracting facial position variation information are not robust. For example, landmark-based methods track the integer coordinates of specific facial landmarks frame-by-frame, which may introduce systematic errors into the sequence since the displacement between frames is usually less than one pixel. Methods based on the Lucas-Kanade algorithm select corner points with good tracking characteristics in the first frame and track the optical flow~\cite{lucas1981iterative} of these points across the video stream, which may cause optical flow disappearance and tracking point drift since the tracking characteristics change with the movement, and the tracking error accumulates.

To make extraction more robust, we propose using virtual anchors to represent the displacement of facial regions. This will eliminate the effects of singular values and noise on tracking a single point. We also propose screening feature points with excellent tracking characteristics frame-by-frame to prevent error accumulation and optical flow disappearance. An overview of our proposed virtual-anchor-based facial region displacement trajectory extraction method is shown in Figure~\ref{fig:fig3}.

\begin{figure}[t]
	\begin{center}
		\includegraphics[width=0.95\linewidth]{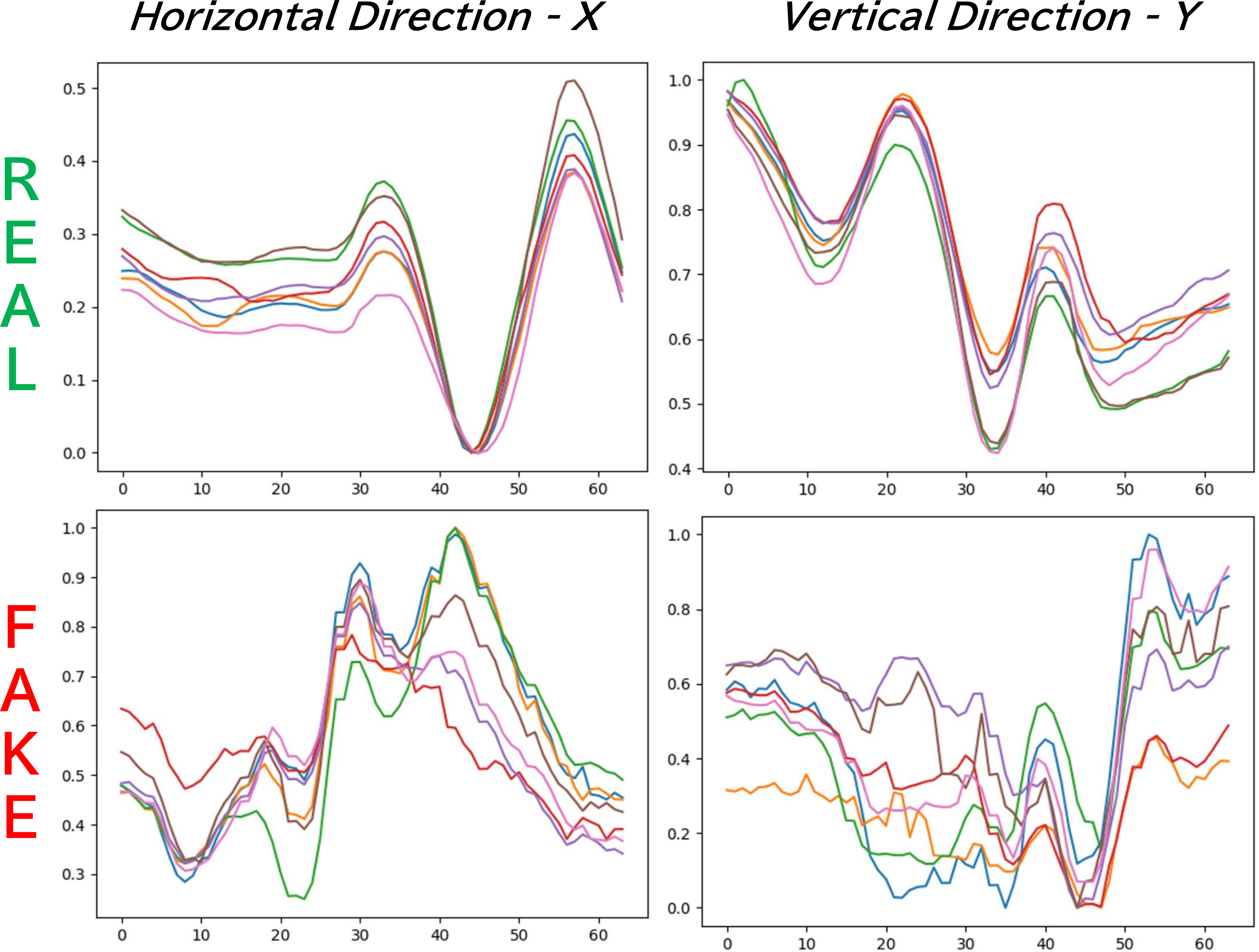}
	\end{center}
	\caption{Example real trajectory (upper row) and example fake trajectory (lower row).}
	\label{fig:fig4}
\end{figure}

\begin{figure*}
	\begin{center}
		\includegraphics[width=0.83\linewidth]{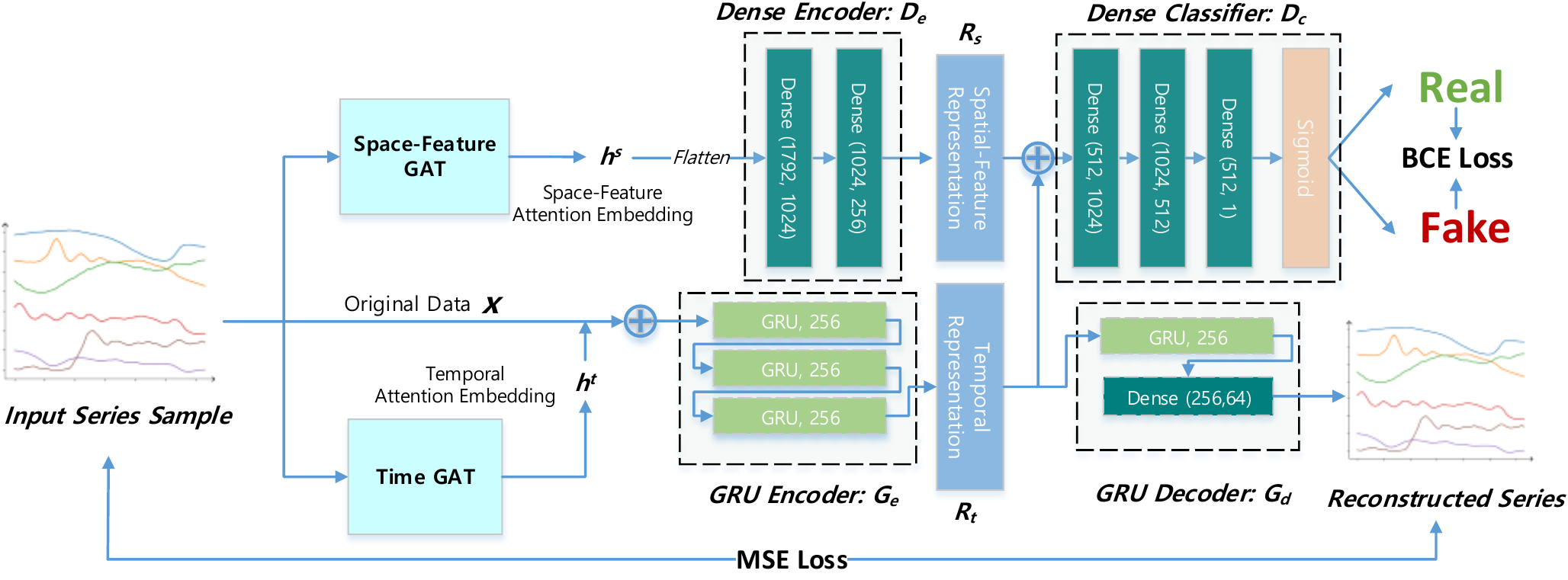}
	\end{center}
	\caption{Architecture of proposed FTDN. Model aggregates dynamic graph attention weights using time and space-feature dimensions of trajectory series and obtains corresponding embeddings. GRU and dense encoders are then used to obtain spatial-temporal representations of data, which are subsequently concatenated and used for deepfakes classification. We also use temporal representation to reconstruct input series to ensure accuracy of GRU encoder in sequence feature learning.}
	\label{fig:fig5}
\end{figure*}

Specifically, for each frame, we first define seven regions of interest (ROIs) in accordance with the landmark coordinates extracted using Dlib~\cite{kazemi2014one}. Since there is no landmark in the forehead area, in order to cope with different face sizes and head poses, we adopt the method in~\cite{niu2020video} to adaptively calculate two feature points for use in dividing the forehead ROI. We then mark tracking points $t$ in each ROI at a uniform interval, which is determined by the width $w$ and height $h$ of the detected face rectangle. The intervals in the X and Y directions are $w/40$ and $h/40$, respectively. Next, for each tracking point $t_i$, we calculate the forward optical flow to obtain position $t^{'}_i$ of the tracking point in the next frame and then calculate the backward optical flow to obtain its tracking offset point $t^{''}_i$. Only the tracking points that are successfully tracked in both the forward and backward optical flows are retained. 

The FB error for $t_i$,
\begin{equation}
FB\_Error_i = |t_i - t^{''}_i|
\end{equation}
is defined as the position drift of the tracking point after forward and backward optical flow tracking and shows the tracking characteristics of the point. As explained in section \ref{ablation}), we discard 50\% of the points in each ROI (those with the largest FB errors) and use the statistical displacement (mean and median) values of the remaining points to represent the displacement of the virtual region anchor. 

For each video, we record a total of 28 features (mean and median values in x and y directions for 7 ROIs). Due to the inconsistent video lengths and face sizes in datasets, we normalize the trajectories and take 64 frames as the window length and 1 s as the sliding length to divide the video trajectory into several sample series with fixed sizes of $28 \times 64$. As shown in Figure~\ref{fig:fig4}, real and fake trajectories can be easily distinguished by the naked eye. For convenience of viewing, we separate the x and y directions and show only the anchor trajectories as calculated using the mean values. Compared with the fake trajectories, the real ones are smoother, and the regional displacements are relatively consistent among ROIs. Moreover, the relative positions between different ROIs for the real trajectory series are basically consistent.

\section{Fake Trajectory Detection Network}
\label{sec:5}
Intuitively, the facial region trajectory samples can be regarded as an interdependent multivariate time series, where the spatial regions and displacement directions are the variables of the series. Therefore, we can transform the problem of deepfakes video detection into how to mine the spatial-temporal features from the multi-variable time series to distinguish inconsistent patterns in the real and fake samples. To effectively utilize the trajectory series samples to identify the artifacts in forged faces, we use our proposed FTDN. The network architecture is shown in Figure~\ref{fig:fig5}.

Specifically, for input series $\pmb{x}$ with size $28 \times 64$, we first calculate the dynamic graph attention embeddings in the time and space-feature dimensions so that the network can adaptively focus on the correlation between each time step in the trajectory series as well as capture the interaction of spatial information between different facial regions and the coherence in different displacement directions. When calculating the space-feature graph attention embedding, we regard the input series as a set of space-feature sequences $S=\{S_1,S_2,…,S_{28}\},S_i\in \mathbb{R}^{64}$ and take each element $S_i$ in the set as the node vector of the fully connected graph to calculate updated output $\pmb{h}^{s}_{i}$:

\begin{equation*}
e_{ij}^{s}=\pmb{a}_s^T LeakyReLU(\pmb{W}_{s}\pmb{S}_i \oplus \pmb{W}_{s}\pmb{S}_j )
\end{equation*}
\begin{equation*}
\alpha_{ij}^{s}=Softmax_j(e_{ij}^s)
\end{equation*}  
\begin{equation}
\pmb{h}_{i}^s=\sigma(\sum_{j}\alpha_{ij}^{s}\pmb{W}_{s}\pmb{S}_j)
\end{equation}  
where $\sigma$ is a sigmoid function, and symbol “$\oplus$” indicates the concatenation of vectors. For space-feature node vectors, we set learnable linear transform matrix $\pmb{W}_{s} \in \mathbb{R}^{64 \times 64}$ and attention vector $\pmb{a}_s^T \in \mathbb{R}^{128}$ so that the updated output $\pmb{h}_{i}^s \in \mathbb{R}^{64}$ after aggregation. This enables us to obtain the space-feature embedding $\pmb{h}^s \in \mathbb{R}^{28 \times 64}$, consistent with the original data. Similarly, when calculating the time graph attention embedding, we regard the features of each time step in the trajectory series as a node vector of the fully connected graph and get a set of sequences $T=\{T_1,T_2,…,T_{64}\},T_i\in \mathbb{R}^{28}$. To ensure that the output time graph attention embedding $\pmb{h}^t$ has the same size as the original data, we set learnable linear transform matrix $\pmb{W}_{t} \in \mathbb{R}^{28 \times 28}$ and attention vector $\pmb{a}_t^T \in \mathbb{R}^{56}$.
    
Next, we concatenate original series $\pmb{x}$ and time graph attention embedding $\pmb{h}^t$ to form a series group with a size of $56 \times 64$ and send it to a GRU encoder. After capturing and modeling the correlation between each independent time step, the encoder outputs a 256-dimensional time descriptor $\pmb{R}_t$ for temporal feature representation. For space-feature attention embedding $\pmb{h}^s$, we flatten the data and send it to a dense encoder to obtain the 256-dimensional spatial-feature representation $\pmb{R}_s$. Process can be formulated as
\begin{equation*}
\pmb{R}_s = D_{e}(flatten(\pmb{h}^s))
\end{equation*}
\begin{equation}
\pmb{R}_t = G_{e}(\pmb{x} \oplus \pmb{h}^t)
\end{equation}
in which $D_{e}$ and $G_{e}$ represent the dense and GRU encoders, respectively. We do not send $\pmb{h}^s$ with $\pmb{h}^s$ and $\pmb{x}$ to the GRU encoder together since each node vector in the space-feature set does not have invariance in the time dimension. During graph information aggregation, the time independence of the reconstructed output is removed, which creates unnecessary redundancies for temporal representation learning.

Finally, we concatenate the two potential representations and input them into the fully connected layers and sigmoid activation. We use binary cross entropy (BCE) as classification loss for real-fake binary classification:

\begin{equation}
L_{BCE} = BCE(D_c(\pmb{R}_s \oplus \pmb{R}_t), \hat{y})
\end{equation}  
where $D_c$ is the dense classifier, and $\hat{y}=\{0,1\}$ is the attribute label of the input series sample. Furthermore, to enhance the modeling of the time descriptor by the GRU encoder, we added an auxiliary learning task. The temporal representation encoded by the GRU encoder is sent to a GRU decoder to reconstruct the multi-variable time series, and then the mean square error (MSE) is used as the reconstruction loss to restrict the similarity between the reconstructed sequence and the input sequence $\pmb{x}$ so as to improve the accuracy of learning the sequence features by the GRU module:

\begin{equation}
L_{MSE} = MSE(G_d(\pmb{R}_t), \pmb{x})
\end{equation}  
where $G_d$ is a GRU decoder composed of a GRU layer and a fully connected layer. To sum up, the total loss consists of two parts:
\begin{equation}
L=L_{BCE}+L_{MSE}
\end{equation}  

\section{Evaluation}
\subsection{Dataset settings}
To evaluate our proposed FTDN, we mainly used the FaceForensics++ (FF++) dataset~\cite{rossler2019faceforensics++} for training and testing. The FF++ dataset consists of a source video sub-dataset extracted from Youtube and several forged video sub-datasets generated by different manipulation techniques, each sub-dataset contains 1000 manipulated videos. For each video in the dataset, the publisher had given three compression versions: uncompressed (raw), moderately-compressed (c23), and strongly-compressed (c40).

We combined the source videos in FF++ dataset with each of the four kinds of manipulated videos (Deepfakes, FaceSwap, Face2Face and NeuralTextures) to form four sub-datasets for our binary classification tasks. The sub-datasets are referred to as DF, FS, F2F, and NT, respectively. For each sub-dataset, we created a training set, a validation set, and a test set at a ratio of 8:1:1. Since the FaceSwap and NeuralTextures videos were slightly shorter than the real videos, when constructing the FS and NT sub-datasets, we used only the top 80\% of the real videos to ensure balance between positive and negative samples. In addition, we separately selected 100 videos with the same index as the DF test set from the FaceShifter to form an unseen sub-dataset called FSH, which is used to test the cross data domain generalization ability of our method. When generating the trajectory samples, we discarded a frame if the frame was not captured, a face was not detected, or the number of points with successful forward and backward optical flow tracking was less than 20\% of the total number of tracking points. If more than ten frames in a video were discarded, we would separate the clips that can continuously detect the face and use them in the test phase.


\subsection{Accuracy Comparison with Baseline}

\begin{table}[t]
\tiny
\renewcommand\arraystretch{1.2} 
\begin{center}
\resizebox{0.48\textwidth}{!}{%
\begin{tabular}{|c|ccccc|}
\hline
\multirow{2}{*}{\textbf{Baseline}} & \multicolumn{5}{c|}{\textbf{Sub-Datasets}}                                                                   \\ \cline{2-6} 
                                   & \textbf{DF}     & \textbf{FS}     & \textbf{F2F}    & \multicolumn{1}{c|}{\textbf{NT}}     & \textbf{FSH}    \\ \hline
\textbf{FakeCatcher~\cite{ciftci2020fakecatcher}}               & 92.5\%          & 94.5\%          & 93.5\%          & \multicolumn{1}{c|}{79.5\%}          & 63.0\%               \\
\textbf{PPG Cell~\cite{ciftci2020hearts}}                  & 94.5\%          & 93.0\%          & 94.5\%          & \multicolumn{1}{c|}{77.0\%}          & 70.0\%          \\
\textbf{RCN~\cite{sabir2019recurrent}}                       & 97.0\%          & 95.5\%          & 95.5\%          & \multicolumn{1}{c|}{85.5\%}          & 65.5\%               \\
\textbf{MesoNet~\cite{afchar2018mesonet}}                   & 97.0\%          & 94.0\%          & 92.0\%          & \multicolumn{1}{c|}{83.5\%}          & 72.0\%          \\
\textbf{Xception~\cite{chollet2017xception}}                  & 99.0\%          & 97.0\%          & 97.5\%          & \multicolumn{1}{c|}{\textbf{93.0\%}} & 65.5\%          \\
\textbf{Capsule-F~\cite{nguyen2019capsule}}                 & 96.5\%          & 97.5\%          & 96.5\%          & \multicolumn{1}{c|}{89.0\%}          & 68.5\%          \\ 
\textbf{LRNet~\cite{sun2021improving}}                 & 98.5\%          & 98.0\%          & 92.0\%          & \multicolumn{1}{c|}{86.5\%}          & 70.0\%          \\ \hline
\textbf{FTDN(Sample)}              & 99.32\%         & 99.16\%         & 98.14\%         & \multicolumn{1}{c|}{90.66\%}         & 75.13\%               \\
\textbf{FTDN(Video)}               & \textbf{99.5\%} & \textbf{99.5\%} & \textbf{98.5\%} & \multicolumn{1}{c|}{92.5\%}          & \textbf{75.0\%} \\ \hline
\end{tabular}}
\end{center}
\caption{The binary classification accuracy comparison on each sub-dataset, and the generalization comparison on unseen dataset.}
\label{tab:tab1}
\end{table}

We evaluated the accuracy of the proposed FTDN by using seven state-of-the-art deepfakes detection methods as our baselines: FakeCatcher~\cite{ciftci2020fakecatcher},  PPG Cell~\cite{ciftci2020hearts}, recurrent convolutional network (RCN) proposed by Sabir \etal~\cite{sabir2019recurrent}, MesoNet~\cite{afchar2018mesonet}, Xception~\cite{chollet2017xception}, Capsule-Forensics~\cite{nguyen2019capsule} and LRNet~\cite{sun2021improving}. For Xception, the benchmark for the FF++ dataset, we modified the final classification layer of the pre-trained model and fine-tuned the parameters by using the FF++ dataset following the guidance of Rossler \etal. For MesoNet, Capsule-Forensics and LRNet, we used the source code provided by the authors to retrain the model and validate the performance by our sub-datasets. For the other methods, we reproduced their works according to the paper. Xception, MesoNet, and Capsule-Forensics were designed for image-level detection, so we had to made some adjustments to adapt them to our video task. Specifically, for a test video, we extracted an image every ten frames to form a sample set, used the well-trained model to predict all the images in the set, and determined the attributes of the test video by majority voting. Similarly, for our method, since each video was divided into several series samples, we used FTDN in the test phase to predict all series samples and determined the result by majority voting. 

The binary classification accuracies of our proposed FTDN on each sub-dataset are listed in Table~\ref{tab:tab1}, including the sample-level and video-level accuracies. The best result for each sub-dataset is shown in bold. FTDN at the video level had the highest accuracy on the DF, FS, and F2F sub-datasets (99.5\%, 99.5\%, and 98.5\%, respectively) and had accuracy close to that of the FF++ benchmark (Xception), which had the highest accuracy (93.0\%), on the NT sub-dataset. Compared with FakeCatcher and PPG Cell, which use facial color series abnormalities to detect deepfakes, the facial trajectory series we extracted is more specific and has better detection performance. Furthermore, in contrast to the LRNet model, which also exposes deepfakes by the geometric information, the GAT mechanism in our method effectively highlights abnormal fluctuations in temporal and spatial dimensions, making our method have better detection performance on F2F and NT sub-datasets.

To verify the generalization ability of our method, which is still a big challenge for all detectors, we trained FTDN and other baselines with DF training set, and check the performance of these models on unseen FSH sub-dataset. As shown in the last column of Table~\ref{tab:tab1}, our method achieved the best detection accuracy of 75\%. In contrast, FakeCatcher showed the worst performance (i.e. 63\%). Although Xception was better than other baselines in the binary classification task, its generalization ability was not as good as MesoNet (i.e. 72\%) with lightweight parameters. RCN model achieved 65.5\% accuracy in cross data domain detection, slightly lower than Capsule-Forensics (i.e. 68.5\%),PPG Cell and LRNet(i.e. 70\%). 

\subsection{Robustness against video compression}
\begin{figure}[t]
	\begin{center}
		\includegraphics[width=1\linewidth]{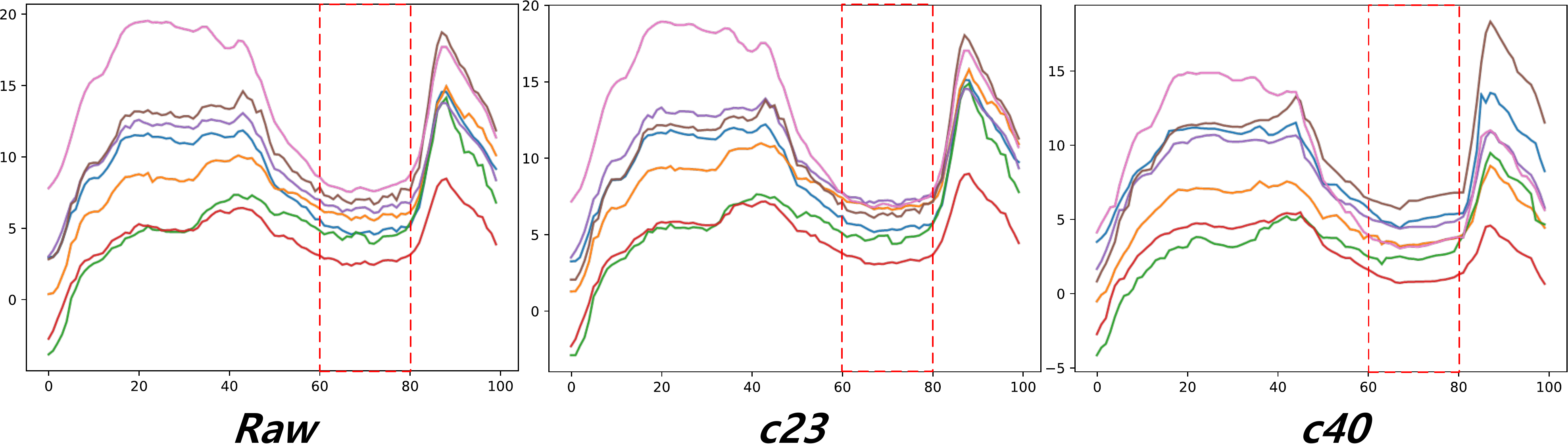}
	\end{center}
	\caption{DeepFake trajectory under different compression levels.}
	\label{fig:fig6}
\end{figure}

Our method models the spatial-temporal features of the displacement trajectory of different facial areas and capture the errors introduced by affine transformation during image blending, which will not be affected by video compression theoretically. However, low image quality has two effects: 1)inaccurate landmark location, which affects the division of ROIs and the feature points involved in tracking; 2)bad video conditions, which add extra noise to the optical flow calculation, which affects the accuracy of feature point tracking. Specific regions of the displacement trajectory under different compression levels are plotted in Figure~\ref{fig:fig6} for the same Deepfakes video. Increasing the degree of compression produced errors in the landmark locations, resulting in a spatial offset in the frame-by-frame ROI division, which changed the relative position of the trajectory. Furthermore, the fuzziness of the pixel information affected the feature point tracking accuracy, which made it difficult to capture the subtle spatial artifacts in local facial areas that resulted from manipulation. This is evidenced by the "smoothing" of the high-frequency burr noise in the areas enclosed by the red box. As noted above, these high-frequency burrs are attributed to the spatial error caused by the forced inverse affine transformation, and are further manifested as the rapid changes in motion trend of local regions during frame-by-frame manipulation, which might provide important evidence for model discrimination.

We tested the anti-compression robustness of our method by using videos in the DF and F2F sub-datasets with different compression levels. The results in Table~\ref{tab:tab2} show that our method maintained excellent performance for moderately-compressed (c23) videos and achieved the highest detection accuracy compared with the baselines. However, for strongly-compressed (c40) videos, its accuracy was severely degraded, and the detection accuracy was lower than that of some image-based methods.


\begin{table}[t]
\tiny
\begin{center}
\renewcommand\arraystretch{1.3} 
\resizebox{0.45\textwidth}{!}{%
\begin{tabular}{|c|cc|cc|}
\hline
\multirow{2}{*}{\textbf{Baseline}} & \multicolumn{2}{c|}{\textbf{DF}} & \multicolumn{2}{c|}{\textbf{F2F}} \\ \cline{2-5} 
                          & \textbf{c23}        & \textbf{c40}        & \textbf{c23}        & \textbf{c40}        \\ \hline
\textbf{PPG cell~\cite{ciftci2020hearts}}                 & 89.43\%    & 79.50\%    &86.41\%          & 77.84\%    \\
\textbf{MesoNet~\cite{afchar2018mesonet}}                  & 93.62\%    & 90.19\%    & 91.98\%     & 80.57\%    \\
\textbf{Xception~\cite{chollet2017xception}}                  & 96.55\%    & \textbf{92.17\%}    & 95.21\%     & 87.39\%    \\
\textbf{Capsule-F~\cite{nguyen2019capsule}}                 & 94.98\%      & 91.06\%    &   94.82\%           & \textbf{89.10\%}           \\ \hline
\textbf{FTDN(ours)}                 & \textbf{98.05\%}    & 83.55\%    & \textbf{96.32\%}     & 83.19\%    \\ \hline
\end{tabular}%
}
\end{center}
\caption{Accuracy comparison on different compression levels.}
\label{tab:tab2}
\end{table}

\begin{figure*}
	\begin{center}
		\includegraphics[width=0.80\linewidth]{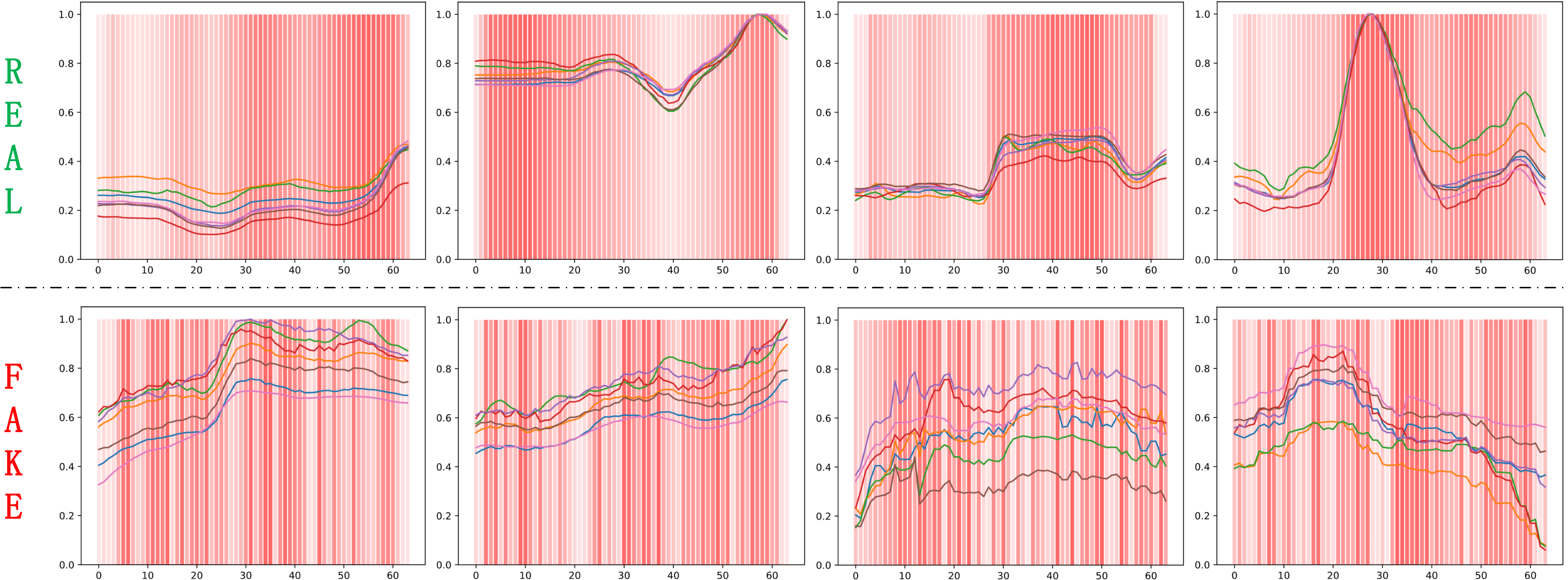}
	\end{center}
	\caption{We use the gradient visualization of real and fake samples to highlight the time steps in the series that contribute significantly to the final classification. The darker the red, the greater impact of the time step on model discrimination.}
	\label{fig:fig7}
\end{figure*}

\subsection{Ablation Study}
\label{ablation}
As mentioned above, when extracting the facial displacement trajectory, we discard 50\% of the tracking points (those with the largest FB errors), and use the mean and median values of the displacement of the remaining tracking points in the region to represent the displacement of the virtual anchor. We investigated the effectiveness without discard, the effectiveness of using the mean value, and the effectiveness of using the median value through an ablation experiment. As shown in Table~\ref{tab:tab3} discarding 50\% of the tracking points with poor tracking characteristics effectively improved detection accuracy. Basically, the effect of using a trajectory based on the mean was better than that of using one based on the median, and a combination of them achieved the best result.

To determine the effectiveness of each component in our network architecture, we conducted ablation experiments on our FTDN, focusing on the effectiveness of the Time GAT (T), the Space-Feature-GAT (SF), and the GRU encoder. The results in Table~\ref{tab:tab4} show that the model relies on the GRU encoder to capture the temporal features of the trajectory series. The Time-GAT and Space-Feature-GAT weight the importance of the sequence information, which helps the GRU improve detection accuracy. Moreover, the contribution of graph attention to F2F and NT is much greater than that to DF and FS. We attribute this to Deepfakes and FaceSwap having a larger area for face swapping, which increases the deviations in their trajectories, enabling the GRU to more easily capture the pattern differences between real and fake samples. In contrast, Face2Face and NeuralTextures affect only subtle areas related to facial expressions, so the effect on the trajectory is relatively minor. The artificial spatial-temporal attention mechanism can thus effectively improve the accuracy of the model.

\begin{table}[t]
\begin{center}
\renewcommand\arraystretch{1.15} 
\resizebox{0.45\textwidth}{!}{%
\begin{tabular}{|c|c|ccc|}
\hline
\textbf{Sub-Dataset} & \textbf{w/o discard} & \textbf{mean+median} & \textbf{mean} & \textbf{median} \\ \hline
\multirow{2}{*}{\textbf{DF}}  & w & \textbf{99.32\%} & 98.70\% & 98.28\% \\
                              & o & 99.03\% & 99.15\% & 97.97\% \\ \hline
\multirow{2}{*}{\textbf{FS}}  & w & \textbf{99.16\%} & 99.07\% & 98.80\%  \\
                              & o & 97.11\% & 97.11\% & 95.89\%   \\ \hline
\multirow{2}{*}{\textbf{F2F}} & w & \textbf{98.14}\% & 96.28\%       &  96.14\%      \\
                              & o & 96.80\% & 95.01\%       & 96.17\%       \\ \hline
\multirow{2}{*}{\textbf{NT}}  & w & \textbf{90.66\%} & 83.34\% & 85.65\% \\
                              & o & 87.37\% & 84.03\% & 86.72\%       \\ \hline
\end{tabular}%
}
\end{center}
\caption{Results of ablation study on facial region displacement trajectory extraction method.}
\label{tab:tab3}
\end{table}

\subsection{Visualization}
Inspired by the concept of gradient-weighted class activation mapping ~\cite{selvaraju2017grad}, we used gradient information to visualize the focus of the model on samples. Figure~\ref{fig:fig7} shows the contribution of trajectory information for each time step to model discrimination (the darker the color, the greater the contribution). The model identifies real samples by identifying continuous, smooth, and stable trajectory patterns, so the time steps focused on for real samples are often successive. In contrast, the model identifies fake samples by identifying the burr noise caused by rapid changes in motion trend and focuses on the time steps in which the relative positions of different ROI sequences change. Therefore, for the fake trajectory series, the region focused on by the model is relatively discrete. 

\begin{table}[t]
\begin{center}
\tiny
\renewcommand\arraystretch{1.15} 
\resizebox{0.45\textwidth}{!}{%
\begin{tabular}{|c|cccc|}
\hline
\multirow{2}{*}{\textbf{Settings}} & \multicolumn{4}{c|}{\textbf{Sub-Dataset}}              \\ \cline{2-5} 
                                   & \textbf{DF} & \textbf{FS} & \textbf{F2F} & \textbf{NT} \\ \hline
\textbf{FTDN}                      & 99.32\%      & 99.16\%      & 98.14\%       & 90.66\%      \\ \hline
\textbf{w/o T}                    & 98.68\%      & 97.65\%      & 93.55\%       & 84.81\%      \\ \hline
\textbf{w/o SF}                   & 98.90\%      & 97.50\%      & 96.21\%       & 88.79\%      \\ \hline
\textbf{w/o T and SF}                 & 98.72\%      & 97.40\%      & 93.55\%       & 82.11\%      \\ \hline
\textbf{w/o GRU}                  & 52.80\%      & 49.32\%      & 56.77\%       & 51.12\%      \\ \hline
\end{tabular}%
}
\end{center}
\caption{Results of ablation study on FTDN model.}
\label{tab:tab4}
\end{table}

\section{Conclusion}
We have devised a deepfakes detection method based on the facial region displacement trajectory. Specifically, we propose using virtual-anchor-based region displacement trajectory extraction to obtain the spatial-temporal representation of different facial areas robustly and accurately. We have also constructed a fake trajectory detection network based on dual-stream spatial-temporal graph attention and a gated recurrent unit backbone that converts the deepfakes detection task into a binary classification problem for a multi-variable time series. Our detection method exhibited competitive performance on samples from the FaceForensics++ dataset.

\section*{Acknowledgements}

This work was partially supported by JSPS KAKENHI Grants JP16H06302, JP18H04120, JP20K23355, JP21H04907, and JP21K18023, and by JST CREST Grants JPMJCR18A6 and JPMJCR20D3, including the AIP challenge program, Japan.

{\small
\bibliographystyle{ieee}
\bibliography{egbib}

\begin{thebibliography}{10}\itemsep=-1pt

\bibitem{FaceApp}
Face{A}pp.
\newblock \url{https://faceappdownload.org}, 2017.

\bibitem{ZAO}
Z{AO}.
\newblock \url{https://apps.apple.com/cn/app/id1465199127}, 2019.

\bibitem{afchar2018mesonet}
Darius Afchar, Vincent Nozick, Junichi Yamagishi, and Isao Echizen.
\newblock Mesonet: a compact facial video forgery detection network.
\newblock In {\em 2018 IEEE international workshop on information forensics and
  security (WIFS)}, pages 1--7. IEEE, 2018.

\bibitem{agarwal2019protecting}
Shruti Agarwal, Hany Farid, Yuming Gu, Mingming He, Koki Nagano, and Hao Li.
\newblock Protecting world leaders against deep fakes.
\newblock In {\em CVPR workshops}, volume~1, page~38, 2019.

\bibitem{brody2021attentive}
Shaked Brody, Uri Alon, and Eran Yahav.
\newblock How attentive are graph attention networks?
\newblock {\em arXiv preprint arXiv:2105.14491}, 2021.

\bibitem{bruna2013spectral}
Joan Bruna, Wojciech Zaremba, Arthur Szlam, and Yann LeCun.
\newblock Spectral networks and locally connected networks on graphs.
\newblock {\em arXiv preprint arXiv:1312.6203}, 2013.

\bibitem{cho2014learning}
Kyunghyun Cho, Bart Van~Merri{\"e}nboer, Caglar Gulcehre, Dzmitry Bahdanau,
  Fethi Bougares, Holger Schwenk, and Yoshua Bengio.
\newblock Learning phrase representations using rnn encoder-decoder for
  statistical machine translation.
\newblock {\em arXiv preprint arXiv:1406.1078}, 2014.

\bibitem{chollet2017xception}
Fran{\c{c}}ois Chollet.
\newblock Xception: Deep learning with depthwise separable convolutions.
\newblock In {\em Proceedings of the IEEE conference on computer vision and
  pattern recognition}, pages 1251--1258, 2017.

\bibitem{ciftci2020fakecatcher}
Umur~Aybars Ciftci, Ilke Demir, and Lijun Yin.
\newblock Fakecatcher: Detection of synthetic portrait videos using biological
  signals.
\newblock {\em IEEE Transactions on Pattern Analysis and Machine Intelligence},
  2020.

\bibitem{ciftci2020hearts}
Umur~Aybars Ciftci, Ilke Demir, and Lijun Yin.
\newblock How do the hearts of deep fakes beat? deep fake source detection via
  interpreting residuals with biological signals.
\newblock In {\em 2020 IEEE international joint conference on biometrics
  (IJCB)}, pages 1--10. IEEE, 2020.

\bibitem{damer2018detecting}
Naser Damer, Viola Boller, Yaza Wainakh, Fadi Boutros, Philipp Terh{\"o}rst,
  Andreas Braun, and Arjan Kuijper.
\newblock Detecting face morphing attacks by analyzing the directed distances
  of facial landmarks shifts.
\newblock In {\em German Conference on Pattern Recognition}, pages 518--534.
  Springer, 2018.

\bibitem{deng2021graph}
Ailin Deng and Bryan Hooi.
\newblock Graph neural network-based anomaly detection in multivariate time
  series.
\newblock In {\em Proceedings of the AAAI Conference on Artificial
  Intelligence}, volume~35, pages 4027--4035, 2021.

\bibitem{goodfellow2014generative}
Ian Goodfellow, Jean Pouget-Abadie, Mehdi Mirza, Bing Xu, David Warde-Farley,
  Sherjil Ozair, Aaron Courville, and Yoshua Bengio.
\newblock Generative adversarial nets.
\newblock {\em Advances in neural information processing systems}, 27, 2014.

\bibitem{guera2018deepfake}
David G{\"u}era and Edward~J Delp.
\newblock Deepfake video detection using recurrent neural networks.
\newblock In {\em 2018 15th IEEE international conference on advanced video and
  signal based surveillance (AVSS)}, pages 1--6. IEEE, 2018.

\bibitem{hochreiter1997long}
Sepp Hochreiter and J{\"u}rgen Schmidhuber.
\newblock Long short-term memory.
\newblock {\em Neural computation}, 9(8):1735--1780, 1997.

\bibitem{itakura1975minimum}
Fumitada Itakura.
\newblock Minimum prediction residual principle applied to speech recognition.
\newblock {\em IEEE Transactions on acoustics, speech, and signal processing},
  23(1):67--72, 1975.

\bibitem{karim2019insights}
Fazle Karim, Somshubra Majumdar, and Houshang Darabi.
\newblock Insights into lstm fully convolutional networks for time series
  classification.
\newblock {\em IEEE Access}, 7:67718--67725, 2019.

\bibitem{kate2016using}
Rohit~J Kate.
\newblock Using dynamic time warping distances as features for improved time
  series classification.
\newblock {\em Data Mining and Knowledge Discovery}, 30(2):283--312, 2016.

\bibitem{kazemi2014one}
Vahid Kazemi and Josephine Sullivan.
\newblock One millisecond face alignment with an ensemble of regression trees.
\newblock In {\em Proceedings of the IEEE conference on computer vision and
  pattern recognition}, pages 1867--1874, 2014.

\bibitem{koopman2018detection}
Marissa Koopman, Andrea~Macarulla Rodriguez, and Zeno Geradts.
\newblock Detection of deepfake video manipulation.
\newblock In {\em The 20th Irish machine vision and image processing conference
  (IMVIP)}, pages 133--136, 2018.

\bibitem{korshunov2018deepfakes}
Pavel Korshunov and S{\'e}bastien Marcel.
\newblock Deepfakes: a new threat to face recognition? assessment and
  detection.
\newblock {\em arXiv preprint arXiv:1812.08685}, 2018.

\bibitem{li2018exposing}
Yuezun Li and Siwei Lyu.
\newblock Exposing deepfake videos by detecting face warping artifacts.
\newblock {\em arXiv preprint arXiv:1811.00656}, 2018.

\bibitem{lucas1981iterative}
Bruce~D Lucas, Takeo Kanade, et~al.
\newblock {\em An iterative image registration technique with an application to
  stereo vision}, volume~81.
\newblock Vancouver, 1981.

\bibitem{malhotra2017timenet}
Pankaj Malhotra, Vishnu TV, Lovekesh Vig, Puneet Agarwal, and Gautam Shroff.
\newblock Timenet: Pre-trained deep recurrent neural network for time series
  classification.
\newblock {\em arXiv preprint arXiv:1706.08838}, 2017.

\bibitem{nguyen2019capsule}
Huy~H Nguyen, Junichi Yamagishi, and Isao Echizen.
\newblock Capsule-forensics: Using capsule networks to detect forged images and
  videos.
\newblock In {\em ICASSP 2019-2019 IEEE International Conference on Acoustics,
  Speech and Signal Processing (ICASSP)}, pages 2307--2311. IEEE, 2019.

\bibitem{niu2020video}
Xuesong Niu, Zitong Yu, Hu Han, Xiaobai Li, Shiguang Shan, and Guoying Zhao.
\newblock Video-based remote physiological measurement via cross-verified
  feature disentangling.
\newblock In {\em European Conference on Computer Vision}, pages 295--310.
  Springer, 2020.

\bibitem{rossler2019faceforensics++}
Andreas Rossler, Davide Cozzolino, Luisa Verdoliva, Christian Riess, Justus
  Thies, and Matthias Nie{\ss}ner.
\newblock Faceforensics++: Learning to detect manipulated facial images.
\newblock In {\em Proceedings of the IEEE/CVF international conference on
  computer vision}, pages 1--11, 2019.

\bibitem{sabir2019recurrent}
Ekraam Sabir, Jiaxin Cheng, Ayush Jaiswal, Wael AbdAlmageed, Iacopo Masi, and
  Prem Natarajan.
\newblock Recurrent convolutional strategies for face manipulation detection in
  videos.
\newblock {\em Interfaces (GUI)}, 3(1):80--87, 2019.

\bibitem{scarselli2008graph}
Franco Scarselli, Marco Gori, Ah~Chung Tsoi, Markus Hagenbuchner, and Gabriele
  Monfardini.
\newblock The graph neural network model.
\newblock {\em IEEE transactions on neural networks}, 20(1):61--80, 2008.

\bibitem{selvaraju2017grad}
Ramprasaath~R Selvaraju, Michael Cogswell, Abhishek Das, Ramakrishna Vedantam,
  Devi Parikh, and Dhruv Batra.
\newblock Grad-cam: Visual explanations from deep networks via gradient-based
  localization.
\newblock In {\em Proceedings of the IEEE international conference on computer
  vision}, pages 618--626, 2017.

\bibitem{sun2021improving}
Zekun Sun, Yujie Han, Zeyu Hua, Na Ruan, and Weijia Jia.
\newblock Improving the efficiency and robustness of deepfakes detection
  through precise geometric features.
\newblock In {\em Proceedings of the IEEE/CVF Conference on Computer Vision and
  Pattern Recognition}, pages 3609--3618, 2021.

\bibitem{velivckovic2017graph}
Petar Veli{\v{c}}kovi{\'c}, Guillem Cucurull, Arantxa Casanova, Adriana Romero,
  Pietro Lio, and Yoshua Bengio.
\newblock Graph attention networks.
\newblock {\em arXiv preprint arXiv:1710.10903}, 2017.

\bibitem{yang2019exposing}
Xin Yang, Yuezun Li, and Siwei Lyu.
\newblock Exposing deep fakes using inconsistent head poses.
\newblock In {\em ICASSP 2019-2019 IEEE International Conference on Acoustics,
  Speech and Signal Processing (ICASSP)}, pages 8261--8265. IEEE, 2019.

\bibitem{yang2019exposinggan}
Xin Yang, Yuezun Li, Honggang Qi, and Siwei Lyu.
\newblock Exposing gan-synthesized faces using landmark locations.
\newblock In {\em Proceedings of the ACM Workshop on Information Hiding and
  Multimedia Security}, pages 113--118, 2019.

\bibitem{zhao2020multivariate}
Hang Zhao, Yujing Wang, Juanyong Duan, Congrui Huang, Defu Cao, Yunhai Tong,
  Bixiong Xu, Jing Bai, Jie Tong, and Qi Zhang.
\newblock Multivariate time-series anomaly detection via graph attention
  network.
\newblock In {\em 2020 IEEE International Conference on Data Mining (ICDM)},
  pages 841--850. IEEE, 2020.

\end{thebibliography}
}

\end{document}